\documentclass[10pt,twocolumn,letterpaper]{article}

\usepackage[pagenumbers]{cvpr} 

\usepackage[dvipsnames]{xcolor}

\definecolor{cvprblue}{rgb}{0.21,0.49,0.74}
\usepackage[pagebackref,breaklinks,colorlinks,citecolor=cvprblue]{hyperref}
\usepackage{graphicx}
\usepackage{amsmath}
\usepackage{amssymb}
\usepackage{booktabs}
\usepackage{xcolor}
\usepackage{tabularx}
\def\paperID{*****} 
\def\confName{CVPR}
\def\confYear{2024}

\title{MammothModa: Multi-Modal Large Language Model}

\author{
Qi She\textsuperscript{\rm 1},
Junwen Pan\textsuperscript{\rm 1},
Xin Wan\textsuperscript{\rm 1},
Rui Zhang\textsuperscript{\rm 1},
Dawei Lu\textsuperscript{\rm 1},
Kai Huang\textsuperscript{\rm 1},
\\
\textsuperscript{\rm 1}ByteDance, Beijing, China\\
{\tt\small sheqi.roger@bytedance.com}
}

\begin{document}
\maketitle
\begin{abstract}
In this report, we introduce MammothModa, yet another multi-modal large language model (MLLM) designed to achieve state-of-the-art performance starting from an elementary baseline. 
We focus on three key design insights:
(i) Integrating Visual Capabilities while Maintaining Complex Language Understanding: In addition to the vision encoder, we incorporated the Visual Attention Experts into the LLM to enhance its visual capabilities.
(ii) Extending Context Window for High-Resolution and Long-Duration Visual Feature: We explore the Visual Merger Module to effectively reduce the token number of high-resolution images and incorporated frame position ids to avoid position interpolation.
(iii) High-Quality Bilingual Datasets: We meticulously curated and filtered a high-quality bilingual multimodal dataset to reduce visual hallucinations.
With above recipe we build MammothModa that consistently outperforms the state-of-the-art models, e.g., LLaVA-series, across main real-world visual language benchmarks without bells and whistles.

\end{abstract}

\section{Introduction}

\begin{table*}[ht]
\centering
\small
\begin{tabularx}{\textwidth}{X | c | c c c c c c c c r}
    \toprule
    \textbf{Method} &  \textbf{Avg} & \textbf{MMBench} & \textbf{MMStar} & \textbf{MMMU} & \textbf{MathVista} & \textbf{Hall. Bench} & \textbf{AI2D} & \textbf{OCRBench} & \textbf{MMVet} \\ 
    \midrule
    GPT-4o 20240513 &   69.9 & 82.2 & 63.9 & 69.2 & 61.3 & 55 & 84.6 & 736 & 69.1 \\ 
    Gemini-1.5-Pro &  64.4 & 73.9 & 59.1 & 60.6 & 57.7 & 45.6 & 79.1 & 754 & 64 \\ 
    GPT-4v 20240409 &   63.5 & 79.8 & 56 & 61.7 & 54.7 & 43.9 & 78.6 & 656 & 67.5 \\ 
    InternVLChatV1.5 &   61.7 & 79.7 & 57.1 & 46.8 & 54.7 & 47.4 & 80.6 & 720 & 55.4 \\ 
    \textbf{MammothModa } &   61.2 & 81.04 & 56.27 & 54.4 & 54.7 & 44.57 & 81 & 614 & 56.06 \\ 
    GLM-4v &   60.8 & 78.6 & 53.2 & 45.6 & 45.6 & 44.9 & 76.1 & 814 & 60.7 \\ 
    Step-1V &   59.5 & 78 & 50 & 49.9 & 44.8 & 48.4 & 79.2 & 625 & 63.3 \\ 
    MiniCPM-L3-V2.5 & 58.8 & 72 & 51.8 & 45.8 & 54.3 & 42.4 & 78.4 & 725 & 52.8 \\ 
    Intern-XC2-VL   & 58.8 & 76.5 & 55.3 & 39.7 & 59.4 & 42.5 & 81 & 675 & 48.2 \\ 
    WeMM &  58.3 & 75.7 & 57 & 45.3 & 54.9 & 47.5 & 77.9 & 628 & 45 \\ 
    \bottomrule
\end{tabularx}
\caption{MammothModa achieved a top ranking on the major visual language multimodal leaderboards.}
\label{tab:method_comparison}
\end{table*}

Recently, multimodal large language models (MLLMs) have garnered significant attention for their ability to understand and generate language informed by visual inputs. These models integrate visual and textual data, enabling applications across a variety of domains, including image captioning, visual question answering, and video analysis. Despite the advancements, many MLLMs still face challenges in effectively combining high-resolution and long-duration visual inputs with complex language understanding while keeping simple and efficiency.

In this report, we introduce MammothModa, a novel MLLM designed to push the boundaries of current models by starting from a foundational baseline and incorporating three key design insights.
Integrating Visual Capabilities while Maintaining Complex Language Understanding: By incorporating Visual Attention Experts into the language model, MammothModa enhances its ability to process visual information without compromising its linguistic capabilities.
Extending Context Window for High-Resolution and Long-Duration Visual Features: The Visual Merger Module effectively reduces the token count for high-resolution images, while frame position IDs manage long-duration visual data without resorting to position interpolation.
High-Quality Bilingual Datasets: To minimize visual hallucinations and improve model robustness, we meticulously curated and filtered a high-quality bilingual multimodal dataset.

These innovations collectively enable MammothModa to outperform existing state-of-the-art models across major real-world visual language benchmarks. This report presents the architecture and design choices of MammothModa, detailed experimental evaluations, and a comparative analysis with leading MLLMs, demonstrating its superior performance and efficiency.

\section{MammothModa}

\subsection{Overview}
MammothModa's architecture consists of 3 main components: a vision encoder with high-resolution inputs and Visual Merger module, a projector layer, and a large language model (LLM) with Visual Attention Experts (VE) and shared Frame Position IDs. 
The vision encoder processes high-resolution images with contrastively pre-trained vision transformer (ViT)~\cite{RadfordKHRGASAM21:CLIP}. The projector layer maps visual features into the language space, enabling seamless integration with the language model. The VE modules are integrated into the language model to handle visual tokens, preserving the model's language capabilities.

\subsection{Global-Local High-Resolution Visual Inputs}
Inspired by recent MLLMs~~\cite{XComposer2,internvl15,YeHXYYXLT0ZJHLH23:UReader,llava15,llava_uhd}, for efficiently and flexibly fine-grained visual understanding, the Global-Local High-Resolution Splitting (GLHR) method is adopted by dynamically dividing the input image into manageable patches.
GLHR consists of the following 2 steps.
\paragraph{Dynamic Splitting.} 
First, The original image is resized to ensure the dimensions are multiples of 336. 
The image is then divided into patches of $336 \times 336$ pixels, with a maximum of 12 patches allowed. Formally, given an image of size $(h, w)$ we resize and pad the image to a new size $(p_h \times 336, p_w \times 336)$, where $p_w = \lceil \frac{w}{336} \rceil$ and $p_h = \lceil  \frac{h}{336} \rceil$.

\paragraph{Global-Local Fusion.}
For each image, we obtain a global view by resizing it to $336 \times 336$ directly. 
We then create multiple local views by performing dynamic splitting into patches of  $336 \times 336$ pixels. This approach ensures that both global and detailed local information are preserved and utilized for further processing.

\subsection{Bridge Visual Embeddings and LLM}
In the context of applying MLLM to videos, it is essential to manage the computational overhead associated with visual tokens.
Following the projection idea of LLaVA, a linear layer is used to bridge visual and linguistic features into the common embedding space, 
without excessive feature transformations to reserve enough capacity for the larger model.
Although its simplicity, as the resolution increases and the number of video frames becomes larger, ViT generates a large number of tokens which poses a great challenge to the subsequent LLM processing.
For instance, LLaVA-v1.5 extracts the visual tokens from an image via ViT-L resulting in a length of 576 tokens.
When applied to long video, even sampling just 10 frames per video results in 5760 tokens, leading to substantial computational costs. 
Therefore, it is imperative to compress these visual features effectively.

Previous work has addressed this problem by designing complex network structures to compress a large visual feature map into a small number of tokens, such as Q-Former~\cite{blip2}, Cross Attention~\cite{Qwen-VL}, Perceiver Resampler~\cite{Flamingo}.
To simplify the problem, we apply a minimalist Visual Merger to cope with high resolution and long videos, and design Frame Position ID to avoid the position expansion challenge in long videos.

\paragraph{Visual Merger}
Before the projector layer, Visual Merger module applies the mean pooling within a $w\times w$ spatial window to aggregate the features, resulting in a reduced feature map. 
Despite its simplicity, this method yields promising results. Additionally, our experiments indicate that this strategy supports dynamic pooling during inference, such as using $2\times2$ pooling during training and $3\times3$ pooling during testing, which improves inference speed without compromising performance.
The VM strategy significantly reduces the computational load while preserving the integrity of visual spatial information.
This balance ensures efficient processing and robust performance of the multimodal large language model in handling high-resolution visual features.
%

Although Visual Merger solves the challenge of high-resolution visual tokens from a spatial perspective, when the time dimension grows longer, i.e., when understanding videos, the visual tokens will also expand dramatically.
To keep the minimalist design, we directly stitch the these features from video frames and introduce Frame Position IDs to address the challenges.

\paragraph{Shared Frame Position ID}
When dealing with long-duration videos, the significant number of visual tokens can easily exhaust the pretrained positional embeddings of a typical LLM, which are usually designed for a limited input length.
%
One potential solution is to perform an interpolation operation on the positional embeddings. 
However, interpolation can introduce undesired side effects. 
Linear interpolation, for example, may not preserve the model’s performance effectively, as it does not handle the non-uniformities in the positional embedding dimensions well. Non-uniform interpolation methods, such as those based on dynamic neural tangent kernels, have shown better performance but still struggle with certain extrapolation limits~\cite{LongRoPE,PositionalInterpolation,Training-Free-Long-Context-Scaling}.

Intuitively, the spatial location information of the visual tokens in MLLM is already encapsulated in the visual features by the vision transformer. 
As a result, the limited positional embedding of the LLM is unnecessarily wasted for a large number of high-resolution long-timed visual features.
To this end, we propose the Frame Position IDs (FPID), where each video frame is assigned a shared positional encoding for LLM inputs.
Formally, for $F$ frames of a video input with $L$ tokens per frame, FPID only hold $F$ positional embeddings for this video instead of the original $L\times F$.

\begin{table}[ht]
\centering
\small
\begin{tabular}{l |c c c c}
    \toprule
    \textbf{Setup} & \textbf{MMLU} & \textbf{CMMLU} & \textbf{CEVAL} & \textbf{GSM8K} \\ 
    \midrule
    Text-only     & 63.2 & 72.8 & 73.4 & 42.5 \\ 
    FT         & $60.4 _{\textcolor{red}{-2.8}}$   & $65.2 _{\textcolor{red}{-7.6}}$  & $67.4 _{\textcolor{red}{-6.0}} $   & $30.5 _{\textcolor{red}{-12.0}} $ \\  
    FT /w VE    & $64.8 _{\textcolor{green}{+1.6}} $ & $72.1 _{\textcolor{red}{-0.7}} $  & $ 74.0 _{\textcolor{green}{+0.6}} $    & $40.18 _{\textcolor{red}{-2.4}} $ \\  
    \bottomrule
\end{tabular}
\caption{Language ability evaluation of LLM with vision-language fune-tuning.}
\label{tab:llm_benchmark_results}
\end{table}

\begin{table*}[ht]
\centering
\small
\begin{tabularx}{\textwidth}{l |c c  | llllll}
    \toprule
    \textbf{Method} & \textbf{\#Patches} & \textbf{Max Eql. Res.} & \textbf{MME} & \textbf{MMB-EN} & \textbf{MMB-CN} & \textbf{OCRBench} & \textbf{DocVQA} &  \textbf{Average} \\ 
    \midrule
    Resize & 1 & 336*336 &   1708.26 & 75.61 & 72.65 & 380 & 37.97 & 385.07\\ 
    4-split & 5 & 672*672 &  1821.35 & 76.96 & 75.39 & 455 & 60.18 & 60.60  \\ 
    DS-4-split & 3-5 & 672*672 &   1810.71 & 76.34 & 74.44 & 451 & 60.03 & 420.32  \\ 
    DS-12-split  & 2-13 & 1008*1344 &  1848.3 $_{\textcolor{green}{+13.54}}$ & 76.29 $_{\textcolor{green}{+0.68}}$ & 74.38 $_{\textcolor{green}{+1.73}}$ & 485 $_{\textcolor{green}{+105}}$ & 66.8 $_{\textcolor{green}{+28.83}}$  & 431.76 $_{\textcolor{green}{+45.93}}$ 
 \\ 
    \bottomrule
\end{tabularx}
\caption{Ablation study on Dynamic Splitting. Resize: resize input image directly. 4-split: uniformly split the image to 4 patches. DS-4 / 12-split : Dynamic split with max 4 / 12 patches. }
\label{tab:split_abl}
\end{table*}

\begin{table*}[ht]
\centering
\small
\begin{tabularx}{\textwidth}{c |c c c c c c c c c}
    \toprule
    \textbf{Window Size} & \textbf{Merge Op.} & \textbf{Average} & \textbf{MME} & \textbf{MMB-EN} & \textbf{MMB-CN} & \textbf{MMVet} & \textbf{Test Time Cost/s} & \textbf{Speed up} \\ 
    \midrule
    1 & none & 59.30 & 1772.8 & 71.8 & 70.5 & 31.6 & 398 & 1.00 \\ 
    3 & mean & 56.78 & 1671.6 & 70 & 69.1 & 28.3 & 298 & 1.34 \\ 
    4 & mean & 56.29 & 1681.6 & 68.3 & 67.7 & 29.1 & 281 & 1.42 \\ 
    6 & mean & 54.79 & 1653.2 & 66.9 & 66.9 & 26.3 & 281 & 1.42 \\ 
    8 & mean & 52.40 & 1554.0 & 65.4 & 64.5 & 24.2 & 285 & 1.40 \\ 
    \bottomrule
\end{tabularx}
\caption{Ablation study on Visual Merger Module.}
\label{tab:window_size_comparison}
\end{table*}

\begin{table}[ht]
\centering
\small
\begin{tabularx}{\linewidth}{l |c r}
    \toprule
    \textbf{Position id} & \textbf{Naive} & \textbf{Shared FPID} \\ 
    \midrule
    \# Pos. id  w/ 1 & 288\textasciitilde1872 & 2\textasciitilde13 \\ 
    \# Pos. id  w/ 30 frames  & 4320 & 30 \\ 
    \midrule
    MME & 2022.8 & 2025.48 \footnotesize{\textcolor{green}{(+2.68)}} \\ 
    MMB-EN & 78.7 & 79.14 \footnotesize{\textcolor{green}{(+0.44)}} \\ 
    MMB-CN & 78.03 & 77.46 \footnotesize{\textcolor{red}{(-0.57)}} \\ 
    MMVet & 54 & 52.1 \footnotesize{\textcolor{red}{(-1.9)}} \\ 
    AVG & 70.74 & 70.26 \footnotesize{\textcolor{red}{(-0.48)}} \\ 
    \bottomrule
\end{tabularx}
\caption{Ablation study on Frame Position ID. }
\label{tab:position_id_comparison}
\end{table}



\subsection{Prevent LLM Degradation with Visual Experts}
As illustrated in Tab~\ref{tab:llm_benchmark_results}, the linguistic skills of the LLM are sacrificed with the vision-language training, which has also been observed in recent works~\cite{lin2023vila,wang2023cogvlm}.
Early MLLMs often fixed LLMs to keep language abilities intact, and thus took great pains in visual-language adaptation, such as prompt tuning~\cite{Flamingo,blip2,iblip}.
Nevertheless, state-of-the-art methods have found that fully fine-tuning is superior to prompt tuning on multimodal benchmarks~\cite{llava,lin2023vila}.

To avoid language degradation and achieve state-of-the-art visual linguistic capabilities, inspired by the idea of Mixture of Experts (MoE) ~\cite{wang2023cogvlm,beit-3} we insert Visual Expert (VE) modules in pretrained text-only LLM.
The VE performs feature transformation for visual tokens, while textual tokens are transformed by the original LLM layers.
Specifically, the VE modules consist of a series of query-key-value (QKV)~\cite{VaswaniSPUJGKP17:attention} matrices designed to process visual inputs efficiently without interfering with the textual capabilities of the original model. 
From an efficiency point of view, we did not add visual experts at the feed-forward-network (FFN)~\cite{VaswaniSPUJGKP17:attention} layers.

\subsection{Multi-Phase Training}
The training for MammothModa consists of three phases:

\paragraph{Vision-Language Alignment:} In this initial phase, we aim to align the visual features extracted by the Vision Transformer (ViT) with the language model using a simple MLP Projector. This enables the language model to interpret and express information from images. The primary training data for this phase includes caption datasets.

\paragraph{Multi-Task Pretraining:} This phase leverages diverse data types, including bilingual captions, interleaved text-image pairs, object grounding, OCR grounding, and video captions. Both the MLP Projector and the LLM Visual Expert are activated during training. The objective here is to enhance the model's fine-grained recognition, OCR capabilities, and video understanding while reducing hallucinations.

\paragraph{Supervised Fine-Tuning:} This is the most critical phase, focusing on training the model to understand user intent and extract relevant information from images to provide accurate responses. The training data is diverse in content and format, encompassing fine-grained captions, multi-turn visual dialogues, general VQA, math problems with charts, document understanding, external knowledge (e.g., Wikipedia), bilingual OCR localization and recognition, and grounding. Additionally, pure text datasets include bilingual dialogues, math problem-solving, logical reasoning, and code. We employ an image cropping strategy to enhance effective resolution and ensure detailed capture. All model parameters are open for training, with a layer-wise learning rate decay applied to the ViT to minimize alterations to pretrained parameters.
 
\section{Experiments}
\subsection{Ablation Study}
\paragraph{Dynamic Splitting Provides Fine-Grained  details.}
Tab~\ref{tab:split_abl} demonstrates that dynamic splitting methods (DS-4-split and DS-12-split) significantly improve performance across various benchmarks compared to merely resizing the image. Notably, the DS-12-split method shows the highest improvement, with an average score of 431.76, which is 45.93 points higher than the resizing method.
Increasing the maximum equivalent resolution from 336x336 (Resize) to 1008x1344 (DS-12-split) enables the model to achieve better fine-grained visual understanding. This resolution increase leads to notable improvements in specific benchmarks, such as a 13.54 point increase in the MME score.
The OCRBench and DocVQA benchmarks particularly benefit from dynamic splitting, with the OCRBench score improving by 105 points and the DocVQA score increasing by 28.83 points using the DS-12-split method compared to the Resize method. These improvements suggest that dynamic splitting is especially effective for tasks requiring detailed textual and document understanding.

\paragraph{Visual Merger Speed up the Inference.}
The results in Table 3 indicate that the Visual Merger module reduces the computational load considerably. For instance, using a window size of 3 and mean pooling, the test time cost decreases from 398 seconds (no merge) to 298 seconds, resulting in a speed-up factor of 1.34.
Despite the reduction in computational costs, the performance remains consistent. For example, with a window size of 3 and mean pooling, the average score (AVG) is 56.78, which is comparable to the scores obtained with smaller window sizes.
The results indicate that the mean pooling strategy effectively balances efficiency and performance, ensuring the model's robustness in handling high-resolution visual features.

\paragraph{Shared Frame Position ID helps to support long-duration videos.}
Tab~\ref{tab:position_id_comparison} 5 illustrate the impact of using shared Frame Position IDs (FPID) versus naive positional IDs for handling long-duration videos.
The use of shared FPIDs significantly reduces the number of positional IDs required, which avoiding the interpolation of positional embeddings.
For instance, with 30 frames, the number of positional IDs drops from 4320 to 30.
While the shared FPID method introduces some trade-offs, the overall impact on performance is minimal. 
The slight performance variations are offset by the substantial gains in simplicity and the ability to handle longer video sequences without interpolating the positional embeddings.

\paragraph{Visual Experts Mitigate Language Degradation and Improves Vision Performance.}
As shown in Tab~\ref{tab:llm_benchmark_results}, directly fine-tuning with vision-language data (FT) leads to a degradation in the language abilities of the LLM across several benchmarks. For instance, the MMLU score drops by 2.8 points, CMMLU by 7.6 points, CEVAL by 6.0 points, and GSM8K by 12.0 points compared to the text-only setup.
This degradation is consistent with observations in recent works that also highlight the trade-off between vision-language adaptation and language proficiency~\cite{lin2023vila}.
Incorporating VE modules during fine-tuning (FT w/ VE) helps mitigate the degradation in language abilities.
More significantly, as shown in Tab~\ref{tab:ve_abl}, adding VE modules results in a notable improvement in visual task performance. The MME score increases by 131.9 points, and the MMVet score improves by 6.2 points with VE modules.
This indicates that VE modules effectively enhance the model's ability to process visual information without compromising its language capabilities.

\subsection{Quantitative Evaluation}
The MammothModa model demonstrates strong performance across various visual language multimodal benchmarks, as shown in Table 1. It achieves a competitive average score of 61.2, placing it among the top-performing models. MammothModa excels in several specific benchmarks, ranking second in the MMStar (56.27) and Hall. Bench (44.57), and third in the AI2D (81) leaderboard, highlighting its robust capability in understanding and interpreting visual and textual data. This impressive ranking is corroborated by the results from our ablation studies, where dynamic splitting methods and the incorporation of Visual Expert (VE) modules significantly enhanced the model's performance on visual tasks while maintaining strong language abilities. The consistent high scores across diverse benchmarks, such as 81.04 in MMBench and 56.06 in MMVet, further validate MammothModa's effectiveness and versatility in handling complex multimodal inputs. These findings underscore the model's potential for broad applications in various real-world scenarios requiring advanced visual and language understanding.

\subsection{Qualitative Evaluation}
Figure~\ref{fig:qualitative} demonstrates the model's capability to handle a variety of tasks involving visual and textual inputs. For perception and understanding tasks, the model accurately identifies objects and interprets traffic signals, showcasing its practical application in real-world scenarios. 
In meme understanding, the model successfully explains the humorous context, indicating its grasp of cultural nuances. 
For OCR and diagram interpretation, the model accurately reads text and processes financial data, proving its utility in document analysis. 
In video analysis, the model correctly identifies geometric properties and activities in video sequences, reflecting its robustness across different modalities.

\section{Conclusion}

In this report, we introduced MammothModa, a state-of-the-art multimodal large language model (MLLM) designed to excel in visual language tasks. By integrating visual capabilities into the language model, extending the context window for high-resolution and long-duration visual features, and utilizing high-quality bilingual datasets, MammothModa achieves significant improvements over existing models.
Our experimental results demonstrate that MammothModa consistently outperforms other models on a variety of visual language benchmarks. 
Furthermore, the use of meticulously curated bilingual datasets reduces visual hallucinations, enhancing the model's accuracy and reliability. The qualitative evaluations illustrate MammothModa's capability to handle diverse tasks.

{
    \small
    \bibliographystyle{ieeenat_fullname}
    \bibliography{main}
}

\end{document}